\documentclass[conference,a4paper]{IEEEtran}

\usepackage[utf8]{inputenc}
\usepackage[T1]{fontenc}
\usepackage{graphicx}
\usepackage{amsmath,amssymb}
\usepackage[english]{babel}
\usepackage{csquotes}
\usepackage[hidelinks]{hyperref}
\usepackage[capitalise]{cleveref}
\usepackage{xcolor}

\graphicspath{{figures/}}

\begin{document}
\bstctlcite{IEEEexample:BSTcontrol}

\title{
    \vspace{-40pt}
\begin{center}
    \small{\textcolor{gray}{Accepted for Publication at IEEE/ASME International Conference on Advanced Intelligent Mechatronics (AIM) 2026 Workshop \\"Robot Learning for Physical Intelligence: Trends, Challenges and Integration"}}
\end{center}
Learning from Relative Action Credit for Escaping Local Minima}

\author{\IEEEauthorblockN{Giovanni Minelli, Giulio Turrisi, Lorenzo Amatucci,
Victor Barasuol, Claudio Semini}
\IEEEauthorblockA{Istituto Italiano di Tecnologia, Genova, Italy\\
\{giovanni.minelli, giulio.turrisi, lorenzo.amatucci, victor.barasuol, claudio.semini\}@iit.it}}

\maketitle

\begin{abstract}
Reinforcement learning holds the promise of robots that learn manipulation skills from interaction alone, without pretrained priors or demonstration data. In practice, on-policy methods such as PPO routinely stall on long-horizon, contact-rich tasks: dense reward shaping carries the policy through stages, yet stitching them together and discovering sparsely rewarded events rarely succeeds. We trace this failure to the missing action discrimination in state-value credit assignment: a state-value critic judges progress only by episode outcomes, so when the policy converges to a partial solution all outcomes coincide, and improvement stops because no action can prove better than another. We therefore extend PPO with an action conditioned value critic which, can compare the executed action with alternatives sampled at the same state and aid the learning signal. Combined with Generalized Advantage Estimation and a bound on exploration noise, we show how the extended algorithm escapes the plateau and completes precise peg insertion within a training budget where PPO with identical exploration fails.
\end{abstract}

\section{Introduction}\label{sec:intro}
\begin{figure}[t]
  \centering
  \includegraphics[width=1.0\columnwidth]{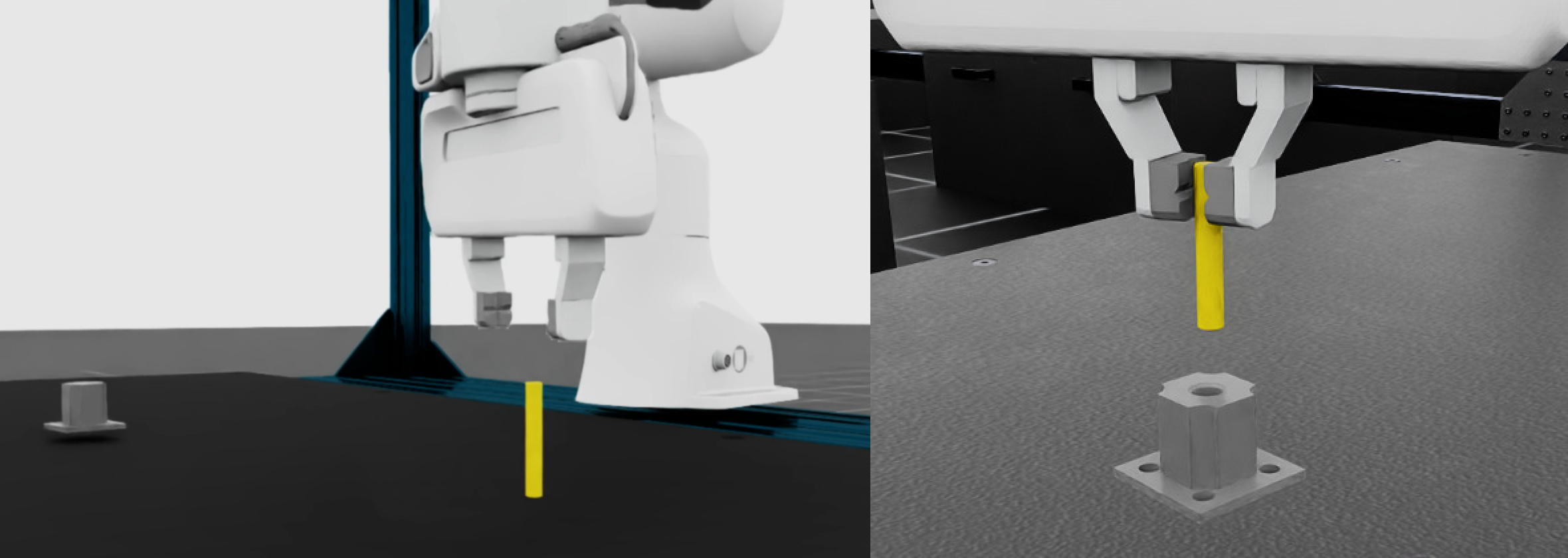}
  \caption{Task scene with detail of the contact-rich insertion stage: the
  peg, grasped and transported, must be aligned with and inserted into a
  tight-fitting hole.}
  \label{fig:task}
\end{figure}

Within the vision of physical AI, a core promise of Reinforcement Learning
(RL) is skill acquisition through interaction alone, without pretrained
models or demonstration data. The de-facto standard realization of this promise is
PPO~\cite{schulman2017ppo} with dense, stage-wise reward shaping in massively
parallel physics simulation~\cite{rudin2022walk}. We study it
on a Factory-based~\cite{narang2022factory} peg-insertion task
(\cref{fig:task}), in which a single policy commanding a 7-DoF arm in joint space must reach the peg, grasp it, lift it, transport and align it over the socket, and finally insert it.

The failure appears consistently, though not always in same phase. PPO learns the first stages of the sequence and then stalls: most often just before insertion; sometimes as early as grasping, where the penalty for knocking the peg over deters further approach; occasionally it completes the
task, but only after a long unproductive plateau. In every case the policy sits in a \emph{shaped-reward local minimum}: it has collected all the intermediate rewards it can reach, and the reward of the next stage lies behind a
low-probability, contact-rich event that Gaussian action noise at the converged policy rarely produces.

% The contributions of this work are: (i) an analysis showing that 
% multi-stage task learning can fail due to collapse of the advantage signal, a structural
% limit of state-value credit assignment rather than a tuning accident;
% (ii) a minimal extension of PPO that adds a group-relative,
% action-conditioned credit signal; and (iii) preliminary simulation results
% in which the extended algorithm escapes the local minimum while PPO, given
% the same exploration, does not.

Prior remedies attack the exploration side: novelty bonuses reward visiting
unfamiliar states~\cite{burda2019rnd}, and maximum-entropy off-policy learning
sustains stochasticity~\cite{haarnoja2018sac}; both add machinery (intrinsic
reward, replay) yet leave credit assignment untouched.
GRPO~\cite{shao2024deepseekmath}, introduced for language-model fine-tuning,
reinforces the responses to a prompt that score above the group mean; we
apply the same principle at every timestep, forming the group from
alternative actions at the same state, scored by a learned critic.

\section{Problem Analysis}\label{sec:failure}
In the task above, the reward is a sum of terms per each
stage (approaching the peg, closing the gripper, gaining height, aligning
to the socket). Once every rollout converges to the same partial solution, two effects compound. First, the state-value critic
$V(s)$ fits this common return and Generalized Advantage Estimation
(GAE)~\cite{schulman2016gae} advantages collapse to near-zero noise. The surrogate gradient loses
direction: nothing in the batch says \enquote{this action was better
than that one}, because at trajectory resolution nothing was. Second,
exploitation pressure shrinks the policy standard deviation faster than the
entropy bonus opposes it. Raising the entropy coefficient is not a straightforward remedy: the
bonus is global, so enough entropy to discover the next stage also corrupts every stage already mastered, a balance ever harder to tune as stages
accumulate. The probability of stumbling through the next
contact-rich event by chance thus decays with the required precision and horizon. The two effects also feed each
other: less exploration means no new reward samples, hence no advantage
signal, hence nothing opposing further exploration decay.

State-value credit assignment also misattributes rare contact events. When
an exploratory action causes a damaging outcome, for example knocking the
peg over, $V$ lowers its estimate of the surrounding states and deters
further search along a path whose actions were mostly right. The longer the
horizon, the wider this deterrence propagates.
% Escaping the minimum requires switching
% the credit question from \enquote{was this state good?} to
% \enquote{was this action good, here?}.

\section{Proposed Method}\label{sec:method}
We keep the PPO machinery intact and change only what the critic can express.
The critic becomes the action-value function $Q(s,a)$, i.e., the expected
return obtained by executing action $a$ in state $s$ and following the
policy thereafter~\cite{sutton2018reinforcement}. The state value used for
bootstrapping is recovered by marginalizing over the current policy with $G$
sampled actions $a_i \sim \pi(\cdot\,|\,s_t)$:
\begin{equation}
  V(s_t) = \tfrac{1}{G}\textstyle\sum_{i=1}^{G} Q(s_t, a_i).
  \label{eq:vfromq}
\end{equation}
GAE is computed as usual from \cref{eq:vfromq}, preserving temporal credit
assignment along the trajectory. The same $G$ evaluations additionally yield
a group-relative advantage in the style of GRPO~\cite{shao2024deepseekmath}:
\begin{equation}
  A^{\mathrm{GRPO}}_t = Q(s_t, a_t) - \tfrac{1}{G}\textstyle\sum_{i=1}^{G} Q(s_t, a_i),
  \label{eq:grpo}
\end{equation}
a contrast between the executed action and counterfactual alternative actions at
the same state. The two signals are combined
additively in the PPO surrogate, $A_t = A^{\mathrm{GAE}}_t +
A^{\mathrm{GRPO}}_t$. Because they measure different quantities (temporal
outcome versus local action ranking), they often disagree in sign, and naively using the group mean cancels the
original learning signal; in practice a directional variant, considering
the group best or worst according to the sign of $A^{\mathrm{GAE}}_t$, is
used to reinforce or dampen the signal. Crucially, the group-relative signal stays informative
when GAE flattens: even under homogeneous returns, $Q$ still ranks actions
within a state, so at the reward plateau the policy gradient is carried almost
entirely by the group term.

The critic is trained by plain regression, $\mathcal{L}_Q =
\big(Q(s_t,a_t)-R^{\lambda}_t\big)^2$, where the target is the
$\lambda$-return of the executed action:
\begin{equation}
  R^{\lambda}_t = r_t + \gamma\big[(1-\lambda)\,V(s_{t+1})
    + \lambda\,R^{\lambda}_{t+1}\big],
  \label{eq:qtarget}
\end{equation}
i.e., the rewards actually collected after taking $a_t$, bootstrapped with
the state value from \cref{eq:vfromq}. Finally, the policy standard deviation is clamped to a floor
($\sigma_{\min}=0.25$) after every optimizer step: the group contrast can
only rank what the sampler proposes, so proposal diversity must not
collapse. We use $G=32$ throughout; actor and critic are small three-layer
MLPs. Overall, the technical delta of the resulting algorithm -- PPO+GRPO in the
following -- over a standard PPO implementation is deliberately minimal: one extra critic input,
$G{-}1$ extra critic forward passes per step, and one extra advantage term, at negligible additional compute.
The escape from minima is powered entirely by local action discrimination.
%
% No demonstrations, no oracle or world model guideing the search

\section{Experimental Results}\label{sec:results}
\begin{figure}[t]
  \centering
  \includegraphics[width=\columnwidth]{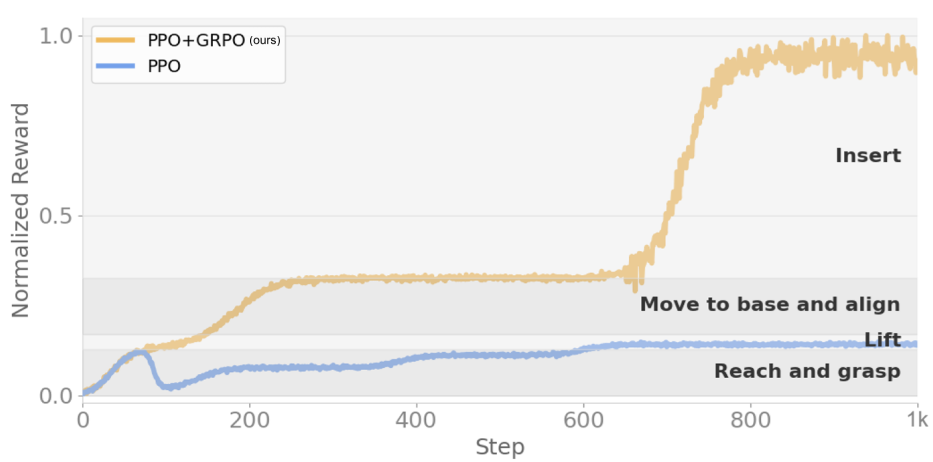}
  \caption{Normalized training reward on the multi-stage peg-insertion task
  (5 seeds); horizontal bands mark the reward level of each completed
  stage. PPO (blue) stalls mid-task without reaching insertion; PPO+GRPO,
  using group-relative action credit (yellow), escapes the plateau and
  converges to consistent insertion.}
  \label{fig:curve}
\end{figure}

\Cref{fig:curve} compares PPO and the proposed method on the peg-insertion
task in GPU-parallel simulation (IsaacLab), averaged over 5
seeds. PPO learns to reach, grasp, and lift, then its return stops improving and
the insertion reward remains zero for the entire run. PPO+GRPO
reaches the same plateau, escapes it after roughly $650$ training
iterations, and converges to about $95\%$ of the maximum return with
consistent insertion.

The exploration floor proves essential: if the standard
deviation is allowed to collapse, all group samples coincide, the $Q$-spread
vanishes, and the contrast degenerates, leaving the action-relative signal
ineffective. On the other hand, granting vanilla PPO the identical
exploration floor does not reliably reproduce the escape: across seeds the
reward stalls, or breaks through only far beyond the budget in which
PPO+GRPO converges. Sustained exploration is therefore
necessary but not sufficient.

A negative result worth reporting: our first working variant adapted the
exploration scale with a CEM-style rule on elite group actions. It escaped
the minimum, but unpredictably, as wide exploratory samples query the critic
out of distribution. Replacing standard deviation adaptation with the standard entropy bonus plus a
hard floor preserved the escape with fewer hyperparameters.
% Useful evaluation metrics for this failure are the iteration at which the escape
% happens and the success rate within a fixed training budget, alongside the curve pair
% that exposes the mechanism: GAE magnitude collapsing while the within-state
% $Q$-spread stays alive.

\section{Conclusion and Future Work}\label{sec:outlook}
We showed that PPO stalls on a multi-stage manipulation task because its
critic credits states, not the executed actions, and that a group-relative action-value
signal restores progress at negligible cost. Both the failure and the
solution connect to broader discussions on continual
learning~\cite{dohare2024plasticity} and on the exploration-exploitation
balance. Ongoing work extends the evaluation to other sequential tabletop
tasks, such as multi-shape peg boards~\cite{luo2024fmb}, and toward real
hardware.
A key open question remains: what learning signal is sufficient to drive
exploration in long-horizon tasks where informative rewards are sparse and
only revealed through world interaction?

\bibliographystyle{IEEEtran}
\bibliography{refs}

\end{document}